\documentclass[12pt]{article}

\usepackage{mslapa}
\usepackage{epsfig}
\usepackage{graphicx}
\usepackage{amsmath}
\usepackage{amsfonts}
\usepackage{bm}
\usepackage{amssymb}
\usepackage{enumerate}
\usepackage{algorithm}
\usepackage{algorithmic}

\newtheorem{theorem}{Theorem}

\newcommand{\bfx}{\mathbf{x}}

\newcommand{\bfw}{\mathbf{w}}

\newcommand{\phib}{\bm{\phi}}
\newcommand{\bfe}{\mathbf{e}}

\begin{document}

\title{Sparse Canonical Correlation Analysis}

\author{David R. Hardoon and John Shawe-Taylor\\
[5pt]Centre for Computational Statistics and Machine Learning\\
Department of Computer Science\\
University College London\\
{\tt \{D.Hardoon,jst\}@cs.ucl.ac.uk }}

\date{}

\maketitle
\begin{abstract}
We present a novel method for solving Canonical Correlation Analysis (CCA) in a sparse convex framework using a least squares approach. The presented method focuses on the scenario when one is interested in (or limited to) a primal representation for the first view while having a dual representation for the second view. Sparse CCA (SCCA) minimises the number of features used in both the primal and dual projections while maximising the correlation between the two views. The method is demonstrated on two paired corpuses of English-French and English-Spanish for mate-retrieval. We are able to observe, in the mate-retreival, that when the number of the original features is large SCCA outperforms Kernel CCA (KCCA), learning the common semantic space from a sparse set of features. 
\end{abstract}

\section{Introduction}
Proposed by \cite{Hotelling36}, CCA is a technique for finding pairs of vectors that maximises the correlation between a set of paired variables. The set of paired variables can be considered as two views of the same object, a perspective we adopt throughout the paper. Since the debut of CCA, a multitude of analyses, adaptations and applications have been proposed \cite{jrketter,fyfe-ica,pei-kernel,akaho01,fclbk01,fblk01,BachJordan,Hardoon03CBMI,Hardoon04,Fukumizu,Hardoon06a,Szedmak07,Hardoon07}.\\
\\
The potential disadvantage of CCA and similar statistical methods, such as Principle Component Analysis (PCA) and Partial Least Squares (PLS), is that the learned projections are a linear combination of all the features in the primal and dual representations respectively. This makes  the interpretation of the solutions difficult. Studies by \cite{zou04sparse,moghaddam-spectral,dhanjal06sparse} and the more recent \cite{Dspremont,sriperumbudurICML07} have addressed this issue for PCA and PLS by learning only the relevant features that maximise the variance for PCA and covariance for PLS. A previous application of sparse CCA has been proposed in \cite{Torres} where the authors imposed sparsity on the semantic space by penalising the cardinality of the solution vector \cite{Weston}. The SCCA presented in this paper is novel to the extent that instead of working with covariance matrices \cite{Torres}, which may be computationally intensive to compute when the dimensionality of the data is large, it deals directly with the training data. \\
\\
In the Machine Learning (ML) community it is common practice to refer to the input space as the primal-representation and the kernel space as the dual-representation. In order to avoid confusion with the meanings of the terms primal and dual commonly used in the optimisation literature, we will use ML-primal to refer to the input space and ML-dual to refer to the kernel space for the remainder of the paper, though note that the references to primal and dual in the abstract refer to ML-primal and ML-dual.\\
\\
We introduce a new convex least squares variant of CCA which seeks a semantic projection that uses as few relevant features as possible to explain as much correlation as possible. In previous studies, CCA had either been formulated in the ML-primal (input) or ML-dual (kernel) representation for both views. These formulations, coupled with the need for sparsity, could prove insufficient when one desires or is limited to a ML primal-dual representation, i.e. one wishes to learn the correlation of words in one language that map to documents in another. We address these possible scenarios by formulating SCCA in a ML primal-dual framework in which one view is represented in the ML-primal and the other in the ML-dual (kernel defined) representation. We compare SCCA with KCCA on a bilingual English-French and English-Spanish data-set for a mate retrieval task. We show that in the mate retrieval task SCCA performs as well as KCCA when the number of original features is small and SCCA outperforms KCCA when the number of original features is large. This emphasises SCCA's ability to learn the semantic space from a small number of relevant features.\\
\\
In Section \ref{sec1} we give a brief review of CCA, and Section \ref{sec2} formulates and defines  SCCA. In Section \ref{sec4} we derive our optimisation problem and show how all the pieces are assembled to give the complete algorithm. The experiments on the paired bilingual data-sets are given in Section \ref{sec5}. Section \ref{sec6} concludes this paper.

\section{Canonical Correlation Analysis}
\label{sec1}
We briefly review canonical correlation analysis and its ML-dual (kernel) variant to provide a smooth understanding of the transition to the sparse formulation. First, basic notation representation used in the paper is defined
\begin{eqnarray*}
\mathbf{b} & - & \text{boldface lower case letters represent vectors}\\
s & - & \text{lower case letters represent scalars}\\
M & - & \text{upper case letters represent matrices}.
\end{eqnarray*}
\def\omit{
Consider the linear combination $x_a = \bfw_a' \mathbf{x}_a$ and $x_b = \bfw_b'\mathbf{x}_b $. Let $\mathbf{x}_a$ and $\mathbf{x}_b$ be two random variables (i.i.d assumption\footnote{The i.i.d assumption is made throughout the paper}) , with zero mean (i.e. the data is centred).\\}
\\
The correlation between $\mathbf{x}_a$ and $\mathbf{x}_b$ can be computed as
\begin{equation}
\label{math:sim}
\max_{\bfw_a,\bfw_b}\rho = \frac{\bfw_a'C_{ab}\bfw_b}{\sqrt{\bfw_a'C_{aa}\bfw_a'\bfw_b'C_{bb}\bfw_b}},
\end{equation}
where $C_{{aa}} =X_aX_a'$ and $C_{{bb}} = X_bX_b'$ are the within-set covariance matrices and $C_{{ab}} = X_aX_b'$ is the between-sets covariance matrix, $X_a$ is the  matrix whose columns are the vectors $\bfx_{i}$, $i = 1,\ldots, \ell$ from the first representation while $X_b$ is the matrix with columns $\bfx_{i}$ from the second representation.  We are able to observe that scaling $\bfw_a, \bfw_b$ does not effect the quotient in equation (\ref{math:sim}), which is therefore equivalent to maximising $\bfw_a'C_{ab}\bfw_b$ subject to $\bfw_a'C_{{aa}}\bfw_a= \bfw_b'C_{{bb}}\bfw_b
=1$.\\
\\
The kernelising of CCA  \cite{fyfe-ica,pei-kernel} offers an alternative by first projecting the data into a higher dimensional feature space  $\phib_t : \bfx = (x_{1},\ldots, x_{n}) \rightarrow \phib_t(\bfx) = (\phib_{1}(\bfx),\ldots,\phib_{N}(\bfx)) \mbox{\ \ $(N \geq n, t = a,b)$}$ before performing CCA in the new feature spaces. The kernel variant of CCA is useful when the correlation is believed to exist in some non linear relationship. Given the kernel functions $\kappa_{a}$ and $\kappa_{b}$ let
$K_{\bm{a}} = X_a' X_a$ and $K_{\bm{b}} = X_b'X_b$ be the linear
kernel matrices corresponding to the two representations of the
data, where $X_a$ is now the  matrix whose columns are the vectors
$\phib_a(\bfx_{i})$, $i = 1,\ldots, \ell$ from the first representation while $X_b$ is the matrix with columns $\phib_b(\bfx_{i})$ from the second representation. The weights $\bfw_a$ and $\bfw_b$ can be expressed as a linear combination of the training examples $\bfw_a = X_a \bm{\alpha}$ and $\bfw_b = X_b \bm{\beta}$.  Substitution into the ML-primal
CCA equation (\ref{math:sim}) gives the optimisation
\begin{equation*}
\max_{\bm{\alpha},\bm{\beta}} \rho =
\frac{\bm{\alpha}' K_{\bf a} K_{\bf b}\bm{\beta}}{\sqrt{\bm{\alpha}' K_{\bf a}^2\bm{\alpha} \bm{\beta}K_{\bf b}^2\bm{\beta}}},
\end{equation*}
which is equivalent to maximising $\bm{\alpha}' K_{\bf a} K_{\bf b}\bm{\beta}$ subject to $\bm{\alpha}' K_{\bf a}^{2}\bm{\alpha} = \bm{\beta}' K_{\bf
b}^{2}\bm{\beta} = 1$. This is the ML-dual form of the CCA optimisation problem given in equation (\ref{math:sim}) which can be cast as a generalised eigenvalue problem and for which the first $k$ generalised eigenvectors can be found efficiently. Both CCA and KCCA can be formulated as symmetric eigenproblems.\\
\\
A variety of theoretical analyses have been presented for CCA \cite{akaho01,BachJordan,Fukumizu,Hardoon04,JohnNelloKM,DRH-JST_08}. A common conclusion of some of these analyses is the need to regularise KCCA. For example the quality of the generalisation of the associated pattern function is shown to be controlled by the sum of the squares of the weight vector norms in \cite{DRH-JST_08}. Although there are advantages in using KCCA, which have been demonstrated in various experiments across the literature, we clarify that when using a linear kernel in both views, regularised KCCA is the same as regularised CCA (since the former and latter are linear). Nonetheless using KCCA with a linear kernel can have advantages over CCA, the most important being speed when the number of features is larger than the number of samples.\footnote{The KCCA toolbox used was from the code section http://academic.davidroihardoon.com/ }

\section{Sparse CCA}
\label{sec2}
The motivation for formulating a ML primal-dual SCCA is largely intuitive when faced with real-world problems combined with the need to understand or interpret the found solutions. Consider the following examples as potential case studies which would require ML primal-dual sparse multivariate analysis methods, such as the one proposed.
\begin{itemize}
\item Enzyme prediction; in this problem one would like to uncover the relationship between the enzyme sequence, or more accurately the sub-sequences within each enzyme sequence that are highly correlated with the possible combination of the enzyme reactants. We would like to find a sparse ML-primal weight representation on the enzyme sequence which correlates highly to sparse ML-dual feature vector of the reactants. This will allow a better understanding of the enzyme structure relationship to reactions.
\item Bilingual analysis;  when learning the semantic relationship between two languages, we may want to understand how one language maps from the word space (ML-primal) to the contextual document (ML-dual) space of another language. In both cases we do not want a complete mapping from all the words to all possible contexts but to be able to extract an interpretable relationship from a sparse word representation from one language to a particular and specific context (or sparse combination of) in the other language.
\item Brain analysis; here, one would be interested in finding a (ML-primal) sparse voxel\footnote{A voxel is a pixel representing the smallest three-dimensional point volume referenced in an fMRI (functional magnetic resonance imaging) image of the brain. It is usually approximately $3$mm$\times3$mm. } activation map to some (ML-dual) non-linear stimulus activation (such as musical sequences, images and various other multidimensional input). The potential ability to find only the relevant voxels in the stimuli would remove the particularly problematic issue of thresholding the full voxel activation maps that are conventionally generated.
\end{itemize}
For the scope of this paper we limit ourselves to experiments with the bilingual texts problems.\\
\\
Throughout the paper we only consider the setting when one is interested in a ML-primal representation for the first view and a ML-dual representation for the second view, although it is easily shown that the given derivations hold for the inverted case (i.e. a ML-dual representation for the first view and a ML-primal representation for the second view) which is therefore omitted.\\
\\
Consider a sample from a pair of random vectors (i.i.d assumptions hold) of the form $(\bfx_a^i,\bfx_b^i)$ each with zero mean (i.e. centred) where $i = 1,\ldots, \ell$. Let $X_a$ and $X_b$ be matrices whose columns are the corresponding training samples and let $K_b = X_b'X_b$ be the kernel matrix of the second view and $\bfw_b$ be expressed as a linear combination of the training examples $\bfw_b = X_b\bfe$ (note that $\bfe$ is a general vector and should not be confused with notation sometimes used for unit coordinate vectors). The primal-dual CCA problem can be expressed as a primal-dual Rayleigh quotient
\begin{eqnarray}
\nonumber
\rho & = & \max_{\bfw_a,\bfw_b}\frac{\bfw_a'X_aX_b'\bfw_b}{\sqrt{\bfw_a'X_aX_a'\bfw_a\bfw_bX_bX_b'\bfw_b}}\\
\nonumber
& =& \max_{\bfw_a,\mathbf{e}}\frac{\bfw_a'X_aX_b'X_b \bfe}{\sqrt{\bfw_a'X_aX_a'\bfw_a\bfe'X_b'X_bX_b'X_b\bfe}}\\
&= & \max_{\bfw_a,\mathbf{e}}\frac{\bfw_a'X_aK_b\bfe}{\sqrt{\bfw_a'X_aX_a'\bfw_a\bfe'K_b^2\bfe}},
\label{eig:origin}
\end{eqnarray}
where we choose the primal weights $\bfw_a$ of the first representation and dual features $\bfe$ of the second representation such that the correlation $\rho$ between the two vectors is maximised. As we are able to scale $\bfw_a$ and $\bfe$ without changing the quotient, the maximisation in equation (\ref{eig:origin}) is equal to maximising $\bfw_a'X_aK_b\bfe$ subject to $\bfw_a'X_a'X_a\bfw_a = \bfe'K_b^2\bfe = 1$. For simplicity let $X = X_a$, $\bfw = \bfw_a$ and $K = K_b$. \\
\\
Having provided the initial primal-dual framework we proceed to reformulate the problem as a convex sparse least squares optimisation problem. We are able to show that maximising the correlation between the two vectors $K\bfe$ and $X'\bfw$ can be viewed as minimising the angle between them. Since the angle is invariant to rescaling, we can fix the scaling of one vector and then minimise the norm\footnote{We define $\| \cdot \|$ to be the $2-$norm.} between the two vectors
\begin{equation}
\label{eigsec}
\min_{\bfw,\bfe}\|X'\bfw - K\bfe\|^2
\end{equation}
subject to $\|K\bfe\|^2 = 1$. This intuition is formulated in the following theorem,
\begin{theorem}
\label{theo1}
Vectors $\bfw,\bfe$ are an optimal solution of equation (\ref{eig:origin}) if and only if there exist $\mu,\gamma$ such that $\mu\bfw, \gamma\bfe$ are an optimal solution of equation (\ref{eigsec}).
\end{theorem}
Theorem \ref{theo1} is well known in the statistics community and corresponds to the equivalence between one form of Alternating Conditional Expectation (ACE) and CCA \cite{Breiman,Hastie}.
For an exact proof see Theorem $5.1$ on page $590$ in \cite{Breiman}.\\
\\
Constraining the $2-$norm of $K\bfe$ (or $X'\bfw$) will result in a non convex problem, i.e we will not obtain a positive/negative-definite Hessian matrix. Motivated by the Rayleigh quotient solution for optimising CCA, whose resulting symmetric eigenproblem does {\it not} enforce the $\|K\bfe\|^2 = 1$ constraint, i.e. the optimal solution is invariant to rescaling of the solutions. Therefore we replace the scaling of $\|K\bfe\|^2 = 1$ with the scaling of $\bfe$ to be $\|\bfe\|_\infty = 1$. We will address the resulting convexity when we achieve the final formulation.\\
\\
After finding an optimal CCA solution, we are able to re-normalise $\bfe$ so that $\|K\bfe\|^2 = 1$ holds.  We emphasis that even though $K$ has been removed from the constraint the link to kernels (kernel tricks and RKHS) is represented in the choice of kernel $K$ used for the dual-view, otherwise the presented method is a sparse linear CCA\footnote{One should keep in mind that even kernel CCA is still linear CCA performed in kernel defined feature space}. We can now focus on obtaining an optimal sparse solution for $\bfw,\bfe$.\\
\\
It is obvious that when starting with $\bfw = \bfe = {\bf 0}$ further minimising is impossible. To avoid this trivial solution and to ensure that the constraints hold in our starting condition\footnote{$\|\mathbf{e}\|_\infty = \max(|e_1|,\ldots,|e_\ell|) = 1$, therefore there must be at least one $e_i$ for some $i$ that is equal to $1$.} we set $\|\bfe\|_\infty = 1$ by fixing $e_k = 1$ for some fixed index $1 \leq k \leq \ell$ so that  $\bfe = [e_1,\ldots,e_{k-1},e_{k},e_{k+1},\ldots,e_\ell]$. To further obtain a sparse solution on $\bfe$ we constrain the $1-$norm of the remaining coefficients $\|\tilde{\bfe}\|_1$, where we define $\tilde{\bfe} = [e_1,\ldots,e_{k-1},e_{k+1},\ldots,e_\ell]$. The motivation behind isolating a specific $k$ and constraining the $1-$norm of the remaining coefficients, other than ensuring a non-trivial solution, follows the intuition of wanting to find similarities between the samples given some basis for comparison. In the case of documents, this places the chosen document (indexed by $k$) in a semantic context defined by an additional (sparse) set of documents. This captures our previously stated goal of wanting to be able to extract an interpretable relationship from a sparse word representation from one language to a particular and specific context in the other language. The $j \in \mathbb{N}^\ell$ choices of $k$ correspond to the $\bfe_j, \bfw_j$ projection vectors. We discuss the selections of $k$ and the ensuring of orthogonality of the sparse projections in Section \ref{sccaalgsec}.\\
\\
We are also now able to constrain the $1-$norm of $\bfw$ without effecting the convexity of the problem. This gives the final optimisation as
\begin{equation}
\label{eqn:stacca1}
\min_{\bfw,\bfe} \| X'\bfw - K\bfe\|^2 + \mu\|\bfw\|_1 + \gamma \| \tilde{\bfe}\|_1
\end{equation}
subject to $\|\bfe\|_\infty = 1$. The expression $\|X\bfw - K\bfe\|^2$ is quadratic in the variables $\bfw$ and $\bfe$ and is bounded from below $(\geq 0)$ and hence is convex since it can be expressed as $\|X\bfw - K\bfe\|^2 = C + g'\bfw + f'\bfe + [\bfw' \bfe'] H [\bfw' \bfe']'$. If $H$ were not positive definite taking multiple $\mu$ of the eigenvector $v' = [v'_1 v'_2]$ with negative eigenvalue $\lambda$ would give $C + \mu g'v_1 + \mu f'v_2 + \mu^2 \lambda$ creating arbitrarily large negative values. When minimising subject to linear constraints (1-norms are linear) this makes the whole optimisation convex.\\
\\
While equation (\ref{eqn:stacca1}) is similar to Least Absolute Shrinkage and Selection Operator (LASSO) \cite{Tibshirani94}\footnote{Basis Pursuit Denoising \cite{chen99}}, it is not a standard LASSO problem unless $\bfe$ is fixed. The problem in equation (\ref{eqn:stacca1}) could be considered as a double-barreled LASSO where we are trying to find sparse solutions for {\it both} $\bfw,\bfe$.\nocite{Heiler}


\section{Derivation \& Algorithm}
\label{sec4}
We propose a novel method for solving the optimisation problem represented in equation (\ref{eqn:stacca1}), where the suggested algorithm minimises the gap between the primal and dual Lagrangian solutions using a greedy search on $\bfw, \bfe$. The proposed algorithm finds a sparse $\mathbf{w},\mathbf{e}$ vectors, by iteratively solving between the ML primal and dual formulation in turn. We give the proposed algorithm as the following high-level pseudo-code. A more complete description will follow later;
\begin{itemize}
\item Repeat
\begin{enumerate}
\item Use the dual Lagrangian variables to solve the ML-primal variables
\item Check whether all constraints on ML-primal variables hold
\item Use ML-primal variables to solve the dual Lagrangian variables
\item Check whether all dual Lagrangian variable constraints hold
\item Check whether 2. holds, IF not go to 1.
\end{enumerate}
\item End
\end{itemize}
We have yet to address how to determine which elements in $\bfw,\bfe$ are to be non-zero. We will show that from the derivation given in Section \ref{primald} a lower and upper bound is computed. Combining the bound with the constraints provides us with a criterion for selecting the non-zero elements for both $\bfw$ and $\bfe$. The criteria being that only the respective indices which violate the bound and the various constraints need to be updated.\\
\\
We proceed to give the derivation of our problem. The minimisation
\begin{equation*}
\min_{\bfw,\bfe} \| X'\bfw - K\bfe\|^2 + \mu\|\bfw\|_1 + \gamma \| \tilde{\bfe}\|_1
\end{equation*}
subject to $\|\bfe\|_\infty = 1$ can be written as
\begin{equation*}
\bfw'XX'\bfw + \bfe'K^2\bfe - 2\bfw'XK\bfe + \mu\|\bfw\|_1 + \gamma \| \tilde{\bfe}\|_1
\end{equation*}
subject to $\|\bfe\|_\infty = 1$, where $\mu$, $\gamma$ are fixed positive parameters.\\
\\
To simplify our mathematical notation we revert to uniformly using $\bfe$ in place of $\tilde{\bfe}$, as $k$ will be fixed in an outer loop so that the only requirement is that no update will be made for $e_k$, which can be enforced in the actual algorithm. We further emphasis that we are only interested in the positive spectrum of $\bfe$, which again can be easily enforced by updating any $e_i < 0$ to be $e_i = 0$\footnote{We can also easily enforce the $\|\cdot\|_\infty$ constraint by updating any $e_i > 1$ to be $e_i = 1$.}. Therefore we could rewrite the constraint $\|\bfe\|_\infty = 1 $ as $0 < e_i \leq 1, \forall i \in \mathbb{R}^\ell$ \\
\\
We are able to obtain the corresponding Lagrangian
\begin{equation*}
\mathcal{L} =  \bfw'XX'\bfw + \mathbf{e}'K^2\mathbf{e} - 2\bfw'XK\mathbf{e}+ \mu \|\bfw\|_1+ \gamma \mathbf{e}'\mathbf{j}  - {\bm{\beta}}'\mathbf{e} ,
\end{equation*}
subject to
\begin{eqnarray*}
 \bm{\beta} & \geq & \mathbf{0},
\end{eqnarray*}
where $\bm{\beta}$ is the dual Lagrangian variable on $\bfe$ and $\mu,\gamma$ are positive scale factors as discussed in Theorem \ref{theo1} and $\mathbf{j}$ is the all ones vector. We note that as we algorithmically ensure that $\bfe \geq 0$ we are able to write $ \gamma \| \bfe\|_1= \gamma \mathbf{e}'\mathbf{j}$ as $\|\bfe\|_1 := \sum_{i=1}^\ell |e_i|$.\\
\\
The constants $\mu,\gamma$ can also be considered as the hyper-parameters (or regularisation parameters) common in the LASSO literature, controlling the trade-off between the function objective and the level of sparsity. We show that the scale parameters can be treated as a type of dual Lagrangian parameters to provide an underlying automatic determination of sparsity. This potentially sub-optimal setting still obtains very good results and is discussed in Section \ref{sec:hyper}.\\
\\
To simplify the $1-$norm derivation we express $\bfw$ by its positive and negative components\footnote{This means that $\bfw^+/\bfw^-$ will only have the positive/negative values of $\bfw$ and zero elsewhere.} such that $\bfw =  \bfw^+ - \bfw^-$ subject to $\bfw^+,\bfw^- \geq 0$. We limit ourselves to positive entries in $\bfe$ as we expect to align with a positive subset of articles.\\
\\
This allows us to rewrite the Lagrangian as
\begin{eqnarray}
\label{eqnlarg}
\mathcal{L} & = & (\bfw^+ - \bfw^-)'XX'(\bfw^+ - \bfw^-) + \mathbf{e}'K^2\mathbf{e} \\
\nonumber
&&  - 2(\bfw^+ - \bfw^-)'XK\mathbf{e}  -{\bm{\alpha}^-}'\bfw^- -{\bm{\alpha}^+}'\bfw^+
 -{\bm{\beta}}'\mathbf{e}\\
\nonumber
&&  + \gamma (\mathbf{e}'\mathbf{j}) + \mu((\bfw^+ + \bfw^-)'\mathbf{j}).
\end{eqnarray}
The corresponding Lagrangian in equation (\ref{eqnlarg}) is subject to
\begin{eqnarray*}
\bm{\alpha}^+ &\geq & \mathbf{0}\\
 \bm{\alpha}^- &\geq & \mathbf{0}\\
 \bm{\beta} & \geq & \mathbf{0}.
\end{eqnarray*}
The two new dual Lagrangian variables $\bm{\alpha}^+, \bm{\alpha}^-$ are to uphold the positivity constraints on $\bfw^+, \bfw^-$.

\subsection{SCCA Derivation}
\label{primald}
In this section we will show that the constraints on the dual Lagrangian variables will form the criterion for selecting the non-zero elements from $\bfw$ and $\bfe$. First we define further notations used. Given the data matrix ${X}\in \mathbb{R}^{m \times \ell}$ and Kernel matrix ${K}\in\mathbb{R}^{\ell \times \ell}$ as defined in Section \ref{sec2}, we define the following vectors
\begin{eqnarray*}
\bfw^+ & = & \left[w^+_1,\ldots, w^+_m\right]\\
\bfw^- & = & \left[w^-_1,\ldots, w^-_m\right]\\
\bm{\alpha}^+ & = &\left[ \alpha^+_1,\ldots, \alpha^+_m \right]\\
\bm{\alpha}^- & = &\left[ \alpha^-_1,\ldots, \alpha^-_m \right]\\
\bfe & = & \left[e_1,\ldots,e_\ell \right]\\
\bm{\beta} & = & \left[\beta_1,\ldots,\beta_\ell \right].
\end{eqnarray*}
Throughout this section let $i$ be the index of either $\bfw,\bfe$ that needs to be updated. We use the notation $(\cdot)_i$ or $[\cdot]_i$ to refer to the $i$th index within a vector and $(\cdot)_{ii}$ to refer to the $i$th element on the diagonal of a matrix. \\
 \\
\\
Taking derivatives of equation (\ref{eqnlarg}) in respect to $\bfw^+$, $\bfw^-$, $\mathbf{e}$ and equating to zero gives
\begin{eqnarray}
\label{sccdev1}
\frac{\partial\mathcal{L}}{\partial \bfw^+} & = & 2 XX'(\bfw^+ - \bfw^-) - 2 X'K\mathbf{e} - \bm{\alpha}^+ + \mu \mathbf{j}   = \mathbf{0}\\
\nonumber
\frac{\partial\mathcal{L}}{\partial \bfw^-} & = &- 2 XX'(\bfw^+ - \bfw^-) + 2X'K\mathbf{e} - \bm{\alpha}^- + \mu \mathbf{j} = \mathbf{0}\\
\nonumber
\frac{\partial\mathcal{L}}{\partial \mathbf{e}} & = &2 K^2\mathbf{e}- 2KX'\bfw - \bm{\beta} + \gamma' \mathbf{j} = \mathbf{0},
\end{eqnarray}
adding the first two equations gives
\begin{eqnarray*}
\bm{\alpha}^+ & = &  2\mu \mathbf{j} -\bm{\alpha}^-\\
\bm{\alpha}^- & =&  2\mu \mathbf{j} -\bm{\alpha}^+,
\end{eqnarray*}
implying a lower and upper component-wise bound on $\bm{\alpha}^-,\bm{\alpha}^+$ of
\begin{eqnarray*}
\mathbf{0} &\leq& \bm{\alpha}^- \leq 2\mu\mathbf{j}\\
\mathbf{0} &\leq& \bm{\alpha}^+ \leq 2\mu\mathbf{j}.
\end{eqnarray*}
We use the bound on  $\bm{\alpha}$ to indicate which indices of the vector $\bfw$ need to be updated by only updating the $w_i$'s whose corresponding $\alpha_i$ violates the bound. Similarly, we only update $e_i$ that has a corresponding $\beta_i$ value smaller than $0$.\\
\\
We are able to rewrite the derivative with respect to $\bfw^+$ in terms of $\bm{\alpha}^-$
\begin{eqnarray*}
\frac{\partial\mathcal{L}}{\partial \bfw^+} & = & 2 XX'(\bfw^+ - \bfw^-) - 2X'K\mathbf{e} -  2\mu \mathbf{j}  + \bm{\alpha}^- + \mu \mathbf{j} \\
& = & 2 XX'(\bfw^+ - \bfw^-) - 2X'K\mathbf{e} -  \mu \mathbf{j} + \bm{\alpha}^- .
\end{eqnarray*}
We wish to compute the update rule for the selected indices of $\bfw$. Taking the second derivatives of equation (\ref{eqnlarg}) in respect to $\bfw^+$ and $\bfw^-$, gives
\begin{eqnarray*}
\frac{\partial^2\mathcal{L}}{\partial \bfw^{+2}} & = & 2 XX' \\
\frac{\partial^2\mathcal{L}}{\partial \bfw^{-2}} & =  & 2 XX',\\
\end{eqnarray*}
so for the $\mathbf{i}_i$, the unit vector with entry $1$, we have an exact Taylor series expansion $t^+$ and $t^-$ respectively for $w^+_i$ and $w^-_i$ as
\begin{eqnarray*}
\mathcal{\hat{L}}(\bfw^+ + t^+\mathbf{i}_i) & = &\mathcal{L}(\bfw^+) + \frac{\partial\mathcal{L}}{\partial w^+_i} t^+ + \frac{\partial^2\mathcal{L}}{\partial w^+_i} (t^+)^2\\
\mathcal{\hat {L}}(\bfw^- + t^-\mathbf{i}_i) & =&  \mathcal{L}(\bfw^-) + \frac{\partial\mathcal{L}}{\partial w^-_i} t^- + \frac{\partial^2\mathcal{L}}{\partial w^-_i} (t^-)^2
\end{eqnarray*}
giving us the exact update for $w^+_i$ by setting
\begin{eqnarray*}
\frac{\partial\mathcal{\hat {L}}(\bfw^+ + t^+\mathbf{i}_i)}{\partial t^+}  & = & \left(2 XX'(\bfw^+ -\bfw^-) - 2X'K\mathbf{e} -\bm{\alpha}^+ + \mu \mathbf{j} \right)_i + 4 (XX')_{ii}t^+ = 0\\
\Rightarrow t^+  &= &  \frac{1}{4 (XX')_{ii}}\left[ 2X'K\mathbf{e}  - 2 XX'(\bfw^+ - \bfw^-) - \bm{\alpha}^- + \mu \mathbf{j}\right]_i.
\end{eqnarray*}
Therefore the update for $w^+_i$ is $\Delta w^+_i = t^+$. We also compute the exact update for $w^-_i$ as
\begin{eqnarray*}
\frac{\partial\mathcal{\hat {L}}(\bfw^- + t^-\mathbf{i}_i)}{\partial t^-} & = &  \left(- 2 XX'(\bfw^+ - \bfw^-) + 2X'K\mathbf{e} - \bm{\alpha}^- + \mu \mathbf{j} \right)_i + 4 (XX')_{ii}t^- = 0\\
\Rightarrow t^- & =&  -\frac{1}{4(XX')_{ii}}\left[ 2X'K\mathbf{e}  - 2 XX'(\bfw^+ - \bfw^-) - \bm{\alpha}^- + \mu \mathbf{j}\right]_i,
\end{eqnarray*}
so that the update for $w^-_i$ is  $\Delta w^-_i = t^-$.
Recall that $\bfw = (\bfw^+ - \bfw^-)$, hence the update rule for $w_i$ is
\begin{eqnarray*}
\hat{w}_i \leftarrow w_i + (\Delta w^+_i - \Delta w^-_i).
\end{eqnarray*}
Therefore we find that the new value of $w_i$ should be
\begin{eqnarray*}
\hat{w}_i \leftarrow w_i +\frac{1}{2(XX')_{ii}}\left[ 2X'K\mathbf{e}  - 2 XX'\bfw - \bm{\alpha}^- + \mu \mathbf{j}\right]_i.
\end{eqnarray*}
We must also consider the update of $w_i$ when $\alpha_i$ is within the constraints and $w_i \neq 0$, i.e.  previously $\alpha_i$ had violated the constraints triggering the updated of $w_i$ to be non zero. Notice from equation (\ref{sccdev1}) that
\begin{equation*}
2(XX')_{ii}w_i + 2 \sum_{j\neq i}(XX')_{ij}w_j = 2(X'K\mathbf{e})_i - {\alpha}_i + \mu.
\end{equation*}
It is easy to observe that the only component which can change is $2(XX')_{ii}w_i$, therefore as we need to update $w_i$ towards zero. Hence when $w_i > 0$ the absolute value of the update is
\begin{eqnarray*}
2(XX')_{ii}\Delta w_i & =& 2\mu - \alpha_i\\
\Delta w_i &=& \frac{2\mu - \alpha_i}{2(XX')_{ii}}
\end{eqnarray*}
else when $w_i < 0$ then the update is the negation of 
\begin{eqnarray*}
2(XX')_{ii}\Delta w_i & =& 0 - \alpha_i\\
\Delta w_i &=& \frac{-\alpha_i}{2(XX')_{ii}}
\end{eqnarray*}
so that the update rule is $\hat{w}_i \leftarrow w_i - \Delta w_i$. In the updating of $w_i$ we ensure that $w_i,\hat{w}_i$ do not have opposite signs, i.e. we will always stop at zero before updating in any new direction.\\
\\
We continue by taking second derivatives of the Lagrangian in equation (\ref{eqnlarg}) with respect to $\mathbf{e}$, which gives
\begin{eqnarray*}
\frac{\partial^2\mathcal{L}}{\partial \mathbf{e}^2} & = & 2  K^2,
\end{eqnarray*}
so for $\mathbf{i}_i$, the unit vector with entry $1$, we have an exact Taylor series expansion
\begin{equation*}
\mathcal{\hat {L}}({\bfe} + t\mathbf{i}_i)  = \mathcal{L}(\bfe) + \frac{\partial\mathcal{L}}{\partial{e}_i} t + \frac{\partial^2\mathcal{L}}{\partial{e}_i} (t)^2
\end{equation*}
giving us the following update rule for ${e}_i$
\begin{eqnarray*}
\frac{\partial\mathcal{\hat {L}}({\bfe} + t\mathbf{i}_i)}{\partial t} & = &  (2  K^2\mathbf{e}- 2KX'\bfw - \bm{\beta} + \gamma' \mathbf{j} )_i + 4 K^2_{ii}t = 0\\
\Rightarrow t & = & \frac{1}{4 K^2_{ii}}\left[ 2KX'\bfw - 2  K^2\mathbf{e} + \bm{\beta} - \gamma' \mathbf{j}\right]_i,
\end{eqnarray*}
the update for $\bfe$ is $\Delta e_i = t$. The new value of $e_i$ will be
\begin{equation*}
{\hat {e}}_i \leftarrow {e}_i + \frac{1}{4 K^2_{ii}}\left[ 2KX'\bfw - 2  K^2\mathbf{e} + \bm{\beta} - \gamma' \mathbf{j}\right]_i,
\end{equation*}
again ensuring that $0 \leq \hat{e}_i \leq 1$.

\begin{algorithm}[tp]
\scriptsize
\caption{The SCCA algorithm}
\label{alg:scca}
input: Data matrix $\mathbf{X}\in \mathbb{R}^{m \times \ell}$, Kernel matrix $\mathbf{K}\in\mathbb{R}^{\ell \times \ell}$ and the value $k$.
\begin{algorithmic}
\STATE $\%$ Initialisation:
\STATE $\mathbf{w} = \mathbf{0}$, $\bf j = 1$, $\bfe = \mathbf{0}$, $e_k = 1$
\STATE $\mu = \frac{1}{M}\sum_i^{M} |( 2XK\bm{e})_i|$, $\gamma = \frac{1}{\ell} \sum_i^{\ell}| (2K^2\bm{e})_i|$
\STATE $\bm{\alpha}^- = 2XK\mathbf{e} + \mu \mathbf{j}$
\STATE $I = (\bm{\alpha} < \mathbf{0})\ || \ (\bm{\alpha} > 2\mu\mathbf{j})$
\STATE
\REPEAT
\STATE $\%$ Update the found weight values:
\STATE Converge over $\mathbf{w}$ using Algorithm \ref{alg:sccaw}
\STATE
\STATE $\%$ Find the dual values that are to be updated
\STATE ${\bm{\beta}} = 2K^2{\mathbf{e}}-2KX\bfw + \gamma {\bf{j}}$
\STATE $J = ({\bm{\beta}} < \mathbf{0})$
\STATE
\STATE $\%$ Update the found dual projection values
\STATE Converge over $\mathbf{e}$ using Algorithn \ref{alg:sccae}
\STATE
\STATE $\%$ Find the weight values that are to be updated
\STATE $\bm{\alpha}^- = 2XK\mathbf{e} - 2 XX'\mathbf{w} + \mu \mathbf{j}$
\STATE $I = (\bm{\alpha} < \mathbf{0})\ || \ (\bm{\alpha} > 2\mu\mathbf{j})$
\UNTIL{convergence}
\STATE
\STATE $\bfe = \frac{\bfe}{\|K\bfe\|}$, $\bfw = \frac{\bfw}{\|X'\bfw\|}$
\STATE
\end{algorithmic}
{\bf Output}: Feature directions $\mathbf{w,e}$
\end{algorithm}

\begin{algorithm}[tp]
\scriptsize
\caption{The SCCA algorithm - Convergence over $\bfw$}
\label{alg:sccaw}
\begin{algorithmic}
\REPEAT
\FOR{$i = 1$ to length of $I$}
\IF{${\alpha}_{I_i} > 2\mu$}
\STATE ${\alpha}_{I_i} = 2\mu$
\STATE $\hat{w}_{I_i} \leftarrow w_{I_i} + \frac{1}{2(XX')_{I_i,I_i}}\left[ 2(XK\mathbf{e})_{I_i}  - 2 (XX'\bfw)_{I_i} -{\alpha}^-_{I_i}+ \mu \right]$
\ELSIF{${\alpha}_{I_i} < 0$}
\STATE ${\alpha}_{I_i} = 0$
\STATE $\hat{w}_{I_i} \leftarrow w_{I_i} + \frac{1}{2(XX')_{I_i,I_i}}\left[ 2(XK\mathbf{e})_{I_i}  - 2 (XX'\bfw)_{I_i} -{\alpha}^-_{I_i} + \mu \right]$
\ELSE
\IF{$w_{I_i} > 0$}
\STATE $\hat{w}_{I_i} \leftarrow w_{I_i} - \frac{2\mu-{\alpha}_{I_i}}{2(XX')_{I_i,I_i}}$
\ELSIF{$w_{I_i} < 0$}
\STATE $\hat{w}_{I_i} \leftarrow w_{I_i} + \frac{{\alpha}_{I_i}}{2(XX')_{I_i,I_i}}$
\ENDIF
\ENDIF
\IF{$sign(w_{I_{i}}) \neq sign(\hat{w}_{I_i})$}
\STATE $w_{I_i} = 0$
\ELSE
\STATE $w_{I_i} = \hat{w}_{I_i}$
\ENDIF
\ENDFOR
\UNTIL{convergence over $\bfw$}
\end{algorithmic}
\end{algorithm}

\begin{algorithm}[tp]
\scriptsize
\caption{The SCCA algorithm - Convergence over $\bfe$}
\label{alg:sccae}
\begin{algorithmic}
\REPEAT
\FOR{$i =  1$ to length of $J$}
\IF{$J_i \neq k$}
\STATE ${{e}}_{J_i} \leftarrow {{e}}_{J_i} +  \frac{1}{4 K^2_{{J_i}{J_i}}}\left[ 2(KX'\bfw)_{J_i} - 2 ( K^2\mathbf{e})_{J_i}-{\gamma}\right]$
\IF{${{e}}_{J_i} < 0$}
\STATE ${{e}}_{J_i} = 0$
\ELSIF{${{e}}_{J_i} > 1$}
\STATE  ${{e}}_{J_i}= 1$
\ENDIF
\ENDIF
\ENDFOR
\UNTIL{convergence over $\mathbf{e}$}
\end{algorithmic}
\end{algorithm}

\subsection{SCCA Algorithm}
\label{sccaalgsec}
Observe that in the initial condition when $\bfw = \mathbf{0}$ from equations (\ref{sccdev1}) we are able to treat the scale parameters $\mu, \gamma$ as dual Lagrangian variables and set them to
\begin{eqnarray*}
\mu & = & \frac{1}{m}\sum_i^{m} |( 2XK\bm{e})_i|\\
\gamma & =& \frac{1}{\ell} \sum_i^{\ell}| (2K^2\bm{e})_i|.
\end{eqnarray*}
We emphasise that this is to provide an underlying automatic determination of sparsity and may not be the optimal setting although we show in Section \ref{sec:hyper} that this method works well in practice. Combining all the pieces we give the SCCA algorithm as pseudo-code in Algorithm  \ref{alg:scca}, which takes $k$ as a parameter.
In order to choose the optimal value of $k$ we would need to run the algorithm with all values of $k$ and select the one giving the best objective value. This would be chosen as the first feature. \\
\\
To ensure orthogonality of the extracted features \cite{JohnNelloKM} for each $\bfe_j$ and corresponding $\bfw_j$, we compute the residual matrices ${X}_j$, $j=1,\ldots,\ell$ by projecting the columns of the data onto the orthogonal complement of ${X}_j'({X}_j{X}_j'\bfw_j)$, a procedure known as deflation,
\begin{equation*}
{X}_{j+1} = X_j\left(I - \mathbf{u}_j\mathbf{p}_j'\right),
\end{equation*}
where $U$ is a matrix with columns $\mathbf{u}_j = X_jX_j'\bfw_j$ and $P$ is a matrix with columns $\mathbf{p}_j = \frac{X_jX_j'\mathbf{u}_j}{\mathbf{u}_j'X_jX_j'\mathbf{u}_j}$. The extracted projection directions can be computed (following \cite{JohnNelloKM}) as $ U(P'U)^{-1}$. Similarly we deflate for the dual view
\begin{equation*}
{K}_{j+1} = \left(I - \frac{\tau_j\tau_j'}{\tau_j'\tau_j}\right){K}_j\left(I - \frac{\tau_j\tau_j'}{\tau_j'\tau_j}\right),
\end{equation*}
where $\tau_j = K_j'(K_j'\bfe_j)$ and compute the projection directions as $B(T'K B)^{-1}T$ where $B$ is a matrix with columns $K_j\bfe_j$ and $T$ has columns $\tau_j$. The deflation procedure is illustrated in pseudocode in Algorithm \ref{completealgo}, for a detailed review on deflation we refer the reader to \cite{JohnNelloKM}.\\
\\
Checking each value of $k$ at each iteration is computationally impractical. In our experiments we adopt the very simplistic strategy of picking the values of $k$ in numerical order $k = 1,\ldots,\ell$. Clearly, there exists intermediate options of selecting a small subset of values at each stage, running the algorithm for each and selecting the best of this subset. This and other extension of our work will be focused on in future studies.
\begin{algorithm}[tp]
\scriptsize
\caption{The SCCA algorithm with deflation}
 \label{completealgo}
input: Data matrix $\mathbf{X}\in \mathbb{R}^{m \times \ell}$, Kernel matrix $\mathbf{K}\in\mathbb{R}^{\ell \times \ell}$.
\begin{algorithmic}
\STATE
\STATE ${X}_1 = X$, ${K}_1 = K$
\FOR{$j =  1$ to $\ell$}
\STATE $k = j$
\STATE $[\bfe_j,\bfw_j] = $ SCCA\_Algorithm$1(X_j,K_j,k)$
\STATE
\STATE $\tau_j = K_j'(K_j'\bfe_j)$
\STATE $\mathbf{u}_j = X_jX_j'\bfw_j$
\STATE $\mathbf{p}_j = \frac{X_jX_j'\mathbf{u}_j}{\mathbf{u}_j'X_jX_j'\mathbf{u}_j}$
\STATE
\IF{$j < \ell$}
\STATE ${K}_{j+1} = \left(I - \frac{\tau_j\tau_j'}{\tau_j'\tau_j}\right){K}_j\left(I - \frac{\tau_j\tau_j'}{\tau_j'\tau_j}\right)$
\STATE ${X}_{j+1} = X_j\left(I - \mathbf{u}_j\mathbf{p}_j'\right)$
\ENDIF
\ENDFOR
\STATE
\end{algorithmic}
\end{algorithm}

\section{Experiments}
\label{sec5}

In the following experiments we use two paired English-French and English-Spanish corpora. The English-French corpus consists of $300$ samples with $2637$ English features and $2951$ French features while the English-Spanish corpus consists of $1,000$ samples with $40,629$ English features and $57,796$ Spanish features. The features represent the number of words in each language. Both corpora are pre-processed into a Term Frequency Inverse Document Frequency (TFIDF) representation followed by zero-meaning (centring) and normalisation. The linear kernel was used for the dual view. The best test performance for the KCCA regularisation parameter for the paired corpora was found to be $0.03$. We used this value to ensure that KCCA was not at a disadvantage since SCCA had no parameters to tune.

\subsection{Hyperparameter Validation}
\label{sec:hyper}
\begin{figure*}[tbhp]
\begin{center}
\includegraphics[width=0.6\textwidth]{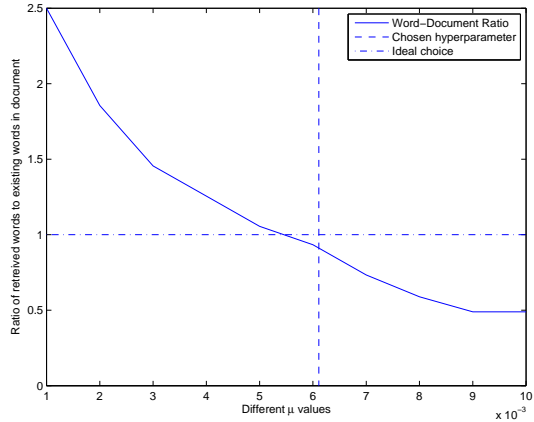}
\end{center}
\caption{Document generation for the English-French corpus (visualisation for a single query): We plot the ratio of total number of selected words to the total number of words in the original document. The horizontal line define the ``ideal choice" where the total number of selected words is identical to the total number of words in the original document. The vertical line represent  the result using the automatic setting of the hyper-parameter. We are able to observe that the automatic selection of $\mu$ is a good approximation for selecting the level of sparsity.}
\label{gif:larschange}
\end{figure*}

In the following section we demonstrate that the proposed approach for automatically determining the regularisation parameter (hyper-parameter) $\mu$ (or alternatively $\gamma$) is sufficient for our purpose. The SCCA problem
\begin{equation}
\label{eqn:stacca}
\min_{\bfw,\bfe} \| X'\bfw - K\bfe\|^2 + \mu\|\bfw\|_1+\gamma\|\tilde{\bfe}\|_1,
\end{equation}
subject to $\|\bfe\|_\infty = 1$ can be simplified to a general LASSO solver by removing the optimisation over $\bfe$, resulting in
\begin{equation*}
\min_{\bfw} \| X'\bfw - \mathbf{k}\|^2 + \mu\|\bfw\|_1,
\end{equation*}
where, given our paired data, ${\bf k}$ is the inner product between the query and the training samples and $X$ is the second paired data samples. This simplified formulation is trivially solved by Algorithm \ref{alg:scca} by ignoring the loops that adapt $\bfe$. The simplification of equation (\ref{eqn:stacca}) allows us to focus on showing that $\mu$ is close to optimal, which is also true for $\gamma$, and therefore omitted.\\
\\
The hyper-parameters control the level of sparsity. Therefore, we test the level of sparsity as a function of the hyper-parameter value. We proceed by creating a new document $d^{*}$ from a paired language that best matches our query\footnote{i.e. given a query in French we want to generate a document in English that best matches the query. The generated document can then be compared to the actual paired English document.} and observe how the change in $\mu$ affects the total number of words being selected. An ``ideal" $\mu$ would generate a new document, in the paired language, and select an equal number of words in the query's actual paired document. Recall that the data has been mean corrected (centred) and therefore no longer sparse.\\
\\
We set $\mu$ to be in the range of $[0.001,\ldots ,1]$ with an increment of $0.001$ and use a leave-paired document-out routine for the English-French corpus, which is repeated for all $300$ documents. Figure \ref{gif:larschange} illustrates, for a single query, the effective change in $\mu$ on the level of sparsity. We plot the ratio of the total number of selected words to the total number of words in the original document. An ideal choice of $\mu$ would choose a ratio of $1$ (the horizontal lines) i.e. create a document with exactly the same number of words as  the original document or in other words select a $\mu$ such that the cross would lie on the plot. We are able to observe that the method for automatically choosing $\mu$ (the vertical line) is able to create a new document with a close approximation to the total number of words in the original document.

In Table \ref{tab:rat} we are able to show that the average ratio of total number of selected words for each document generated in the paired language is very close to the ``ideal" level of sparsity, while a non-sparse method (as expected) generates a document with an average of $\approx 28$ times the number of words from the original document. Now that we have established the automatic setting of the hyper-parameters, we proceed in testing how `good' the selected words are in the form of a mate-retreiveal experiment.
\begin{table}[htdp]
\caption{French-English Corpus: The ratio of the total number of selected words to the actual total number of words in the paired test document, averaged over all queries. The optimal average ratio if we always generate an `ideal' document is $1$.}
\begin{center}
\begin{tabular}{|c|c|}
\hline
 & Average Selection Ratio\\
\hline
Automatic setting of $\mu$  & $1.01\pm0.54$\\
Non-sparse method & $28.15\pm15.71$\\
\hline
\end{tabular}
\end{center}
\label{tab:rat}
\end{table}%

\subsection{Mate Retrieval}
\begin{figure*}[tbhp]
\begin{center}
\includegraphics[width=0.8\textwidth]{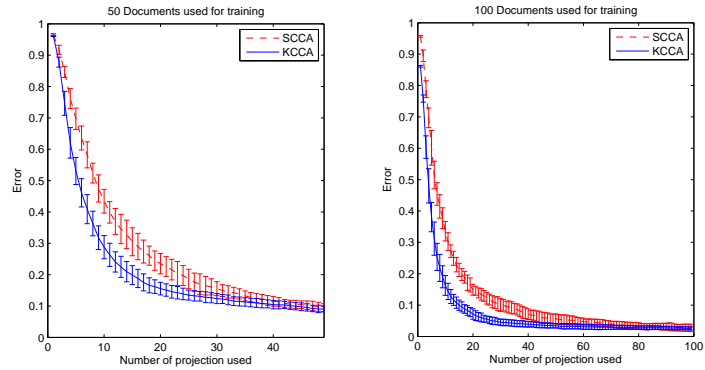}
\end{center}
\caption{English-French: The average precision error (1-$p$) with $\pm$ standard division error bars for  SCCA and KCCA for different number of projections used for the mate-retrieval task. The left figure is for $50$ training and $250$ testing documents while the right figure is for $100$ training and $200$ testing documents.}
\label{fig:EnFr}
\end{figure*}

\begin{figure*}[tbhp]
\begin{center}
\includegraphics[width=0.8\textwidth]{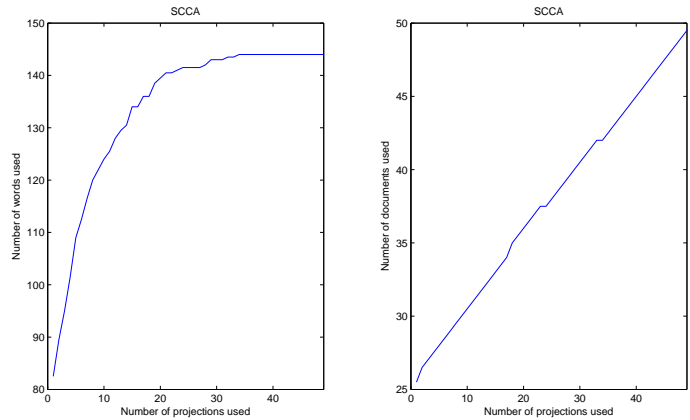}
\end{center}
\caption{English-French: Level of Sparsity - The following figure is an extension of Figure \ref{fig:EnFr} which uses $50$ documents for training. The left figure plots the average number of words used while the right figure plots the average number of documents used with the number of projections. For reference, KCCA uses all the words (average of $2794$) and documents ($50$) for all number of projections.}
\label{fig:EnFrp}
\end{figure*}

\begin{figure*}[tbhp]
\begin{center}
\includegraphics[width=0.7\textwidth]{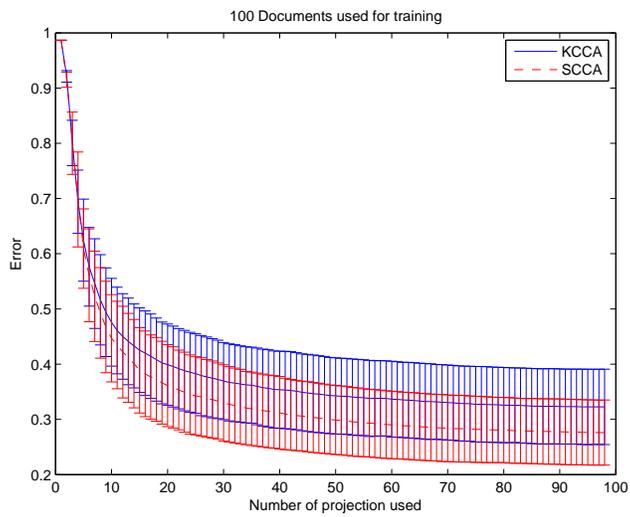}
\end{center}
\caption{English-Spanish: The average precision error (1-$p$) with $\pm$ standard division error bars of SCCA and KCCA for different number of projections used for the mate-retrieval task. We use $100$ documents for training and $900$ for testing documents.}
\label{fig:EnEs}
\end{figure*}

\begin{figure*}[tbhp]
\begin{center}
\includegraphics[width=0.9\textwidth]{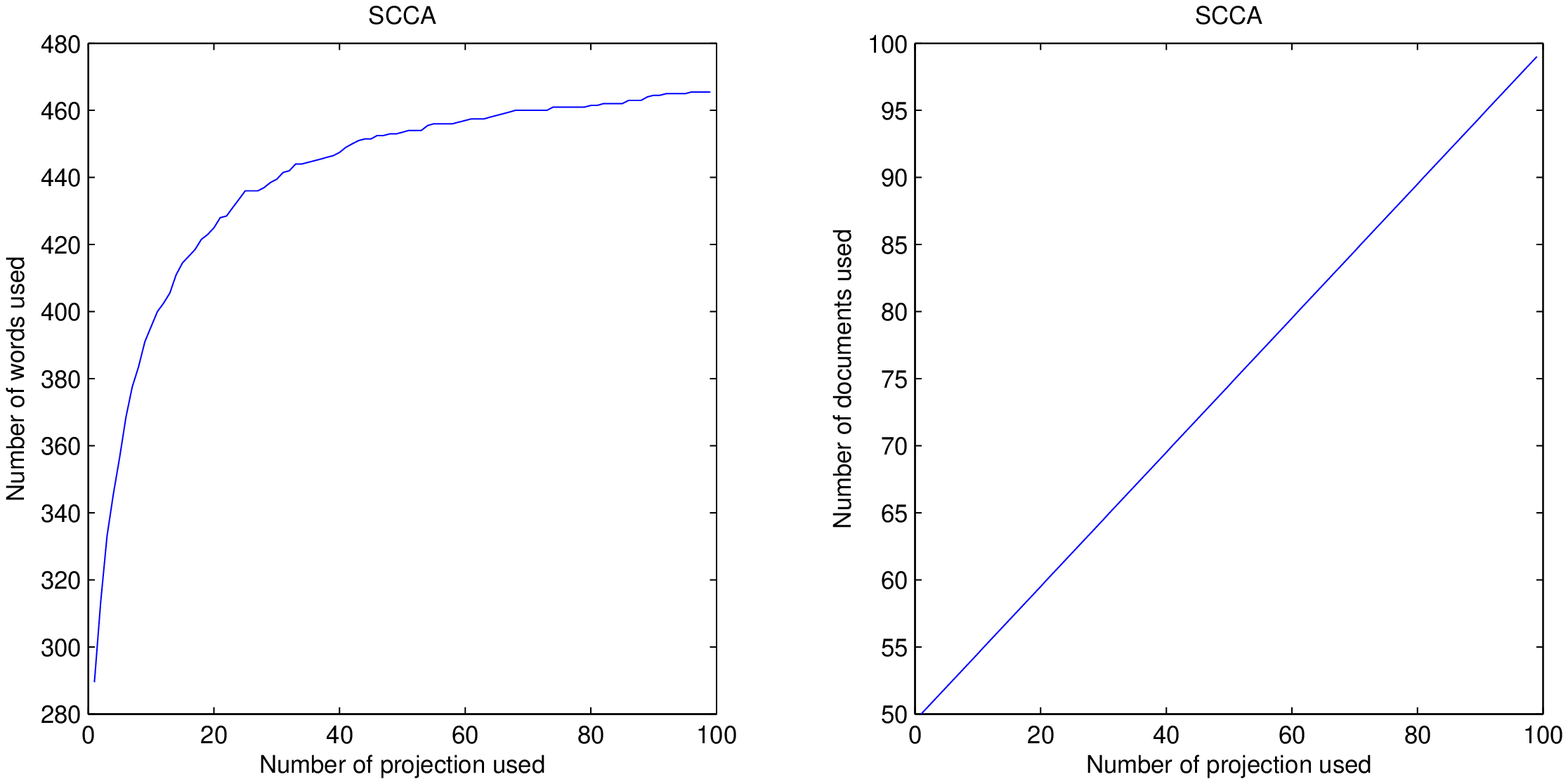}
\end{center}
\caption{English-Spanish: Level of Sparsity - The following figure is an extension of Figure \ref{fig:EnEs} which uses $100$ documents for training. The left figure plots the average number of words used and while the right figure plots the average number of documents used with increasing number of projections. For reference, KCCA uses all the words (average of $49,212$) and documents ($100$) for all number of projections.}
\label{fig:EnEsp}
\end{figure*}

Our experiment is of mate-retrieval, in which a document from the test corpus of one language is considered as the query and only the mate document from the other language is considered relevant. In the following experiments the results are an average of retrieving the mate for both English and French (English and Spanish) and have been repeated $10$ times with a random train-test split.\\
\\
We compute the mate-retrieval by projecting the query document as well as the paired (other language) test documents into the learnt semantic space where the inner product between the projected data is computed. Let $q$ be the query in one language and $K_s$ the kernel matrix of the inner product between the second language's testing and training documents
\begin{equation*}
l = \left< \frac{q'\bfw}{\|q'\bfw\|}, \frac{K_s \bfe}{\|K_s \bfe\|} \right>.
\end{equation*}
The resulting inner products $l$ are then sorted by value. We measure the success of the mate-retrieval task using average precision, this assesses where the correct mate within the sorted inner products $l$ is located. Let $I_j$ be the index location of the retrieved mate from query $q_j$, the average precision $p$ is computed as
\begin{equation*}
p = \frac{1}{M}\sum_{j=1}^M \frac{1}{I_j},
\end{equation*}
where $M$ is the number of query documents.\\
\\
We start by giving the results for the English-French mate-retrieval as shown in Figure \ref{fig:EnFr}. The left plot depicts the average precision ($\pm$ standard deviation) when $50$ documents are used for training and the remaining $250$ are used as test queries. The right plot in Figure \ref{fig:EnFr} gives the average precision ($\pm$ standard deviation) when $100$ documents are used for training and the remaining $200$ for testing. It is interesting to observe that even though SCCA does not learn the common semantic space using all the features (average plotted in Figure \ref{fig:EnFrp}) for either ML primal or dual views (although SCCA will use full dual features when using the full number of projections) its error is extremely similar to that of KCCA and in fact  converges with it when a sufficient number of projections are used. It is important to emphasise that KCCA uses the full number of documents ($50$ and $100$) and the full number of words (an average of $2794$ for both languages) to learn the common semantic space. For example, following the left plot in Figure \ref{fig:EnFr} and the additional plots in Figure \ref{fig:EnFrp} we are able to observe that when $35$ projections are used KCCA and SCCA show a similar error. However, SCCA uses approximately $142$  words and $42$ documents to learn the semantic space, while KCCA uses $2794$ words and $50$ documents.\\
\\
The second mate-retrieval experiment uses the English-Spanish paired corpus. In each run we randomly split the $1000$ samples into $100$ training and $900$ testing paired documents. The results are plotted in Figure \ref{fig:EnEs} where we are clearly able to observe SCCA outperforming KCCA throughout. We believe this to be a good example of when too many features hinder the learnt semantic space, also explaining the difference in the results obtained from the English-French corpus as the number of features are significantly smaller in that case. The average level of SCCA sparsity is plotted in Figure \ref{fig:EnEsp}. In comparison to KCCA which uses all words ($49,212$) SCCA uses a maximum of $460$ words.\\
\\
The performance of SCCA, especially in the latter English-Spanish experiment, shows that we are indeed able to extract meaningful semantics between the two languages, using only the relevant features.


Despite these already impressive results our intuition is that even better results are attainable if the hyper-parameters would be tuned to give optimal results. The question of hyper-parameter optimality is left for future research. Although, it seems that the main gain of SCCA is sparsity and interpretability of the features.

\section{Conclusions}
\label{sec6}
Despite being introduced in $1936$, CCA  has proven to be an inspiration for new and continuing research.  In this paper we analyse the formulation of CCA and address the issues of sparsity as well as convexity by presenting a novel SCCA method formulated as a convex least squares approach. We also provide a different perspective of solving CCA by using a ML primal-dual formulation which focuses on the scenario when one is interested in (or limited to) a ML-primal representation for the first view while having a ML-dual representation for the second view. A greedy optimisation algorithm is derived. \\
\\
The method is demonstrated on a bi-lingual English-French and English-Spanish paired corpora for mate retrieval. The true capacity of SCCA becomes visible when the number of features becomes extremely large as SCCA is able to learn the common semantic space using a very sparse representation of the ML primal-dual views.\\
\\
The papers raison d'\^etre was to propose a new efficient algorithm for solving the sparse CCA problem. We believe that while addressing this problem new and interesting questions which need to be addressed have surfaced
\begin{itemize}
\item How to automatically compute the hyperparameters $\mu,\gamma$ values so to achieve optimal results?
\item How do we set $k$ for each $\bfe_j$ when we wish to compute less than $\ell$ projections?
\item Extending SCCA to a ML primal-primal (ML dual-dual) framework.
\end{itemize}
We believe this work to be an initial stage for a new sparse framework to be explored and extended.


\section*{Acknowledgment}
David R. Hardoon is supported by the EPSRC project Le Strum, EP-D063612-1. We would like to thank Zakria Hussain and Nic Schraudolph for insightful discussions. This publication only reflects the authors views.

\bibliographystyle{mslapa}

\end{document}